\documentclass[11pt,a4paper]{article}
\usepackage{acl2015}
\usepackage{latexsym}
\usepackage{microtype}		
\usepackage{enumitem}		
\usepackage{graphicx}
\usepackage{multirow}
\usepackage{cleveref}
\usepackage{algorithm}
\usepackage{amsmath}
\usepackage{amssymb}
\usepackage{array}
\usepackage{times}
\usepackage{url}

\usepackage{caption}
\usepackage{array}

\usepackage{latexsym}
\usepackage{graphicx}
\usepackage{multicol}
\usepackage{fixltx2e}
\usepackage{enumitem}
\usepackage{subcaption}
\usepackage{subfig}

\title{Multi-Field Structural Decomposition for Question Answering}

\author{Tomasz Jurczyk \\
	    Mathematics and Computer Science\\
	    Emory University\\
	    Atlanta, GA 30322, USA\\
  {\tt tomasz.jurczyk@emory.edu} 
  \And
	   	Jinho D.\ Choi\\
	  	Mathematics and Computer Science\\
	    Emory University\\
	  	Atlanta, GA 30322, USA\\
  {\tt jinho.choi@emory.edu} \\}

\date{}

\begin{document}
\maketitle

\newcounter{LINE_NUM}

\newcommand{\LN}{\addtocounter{LINE_NUM}{1}\arabic{LINE_NUM}:}
\newcommand{\TAB}{$\:\:\:\:$}
\newcommand{\rTRUE}{\texttt{True}}
\newcommand{\rFALSE}{\texttt{False}}
\newcommand{\rNULL}{\texttt{null}}

\newcommand{\LET}[2]{\textbf{let}$\:\:${#1}$\:\:$\textbf{be}$\:\:${#2}}
\newcommand{\SET}[2]{\textbf{set}$\:\:${#1}$\:\:$\textbf{to}$\:\:${#2}}
\newcommand{\uWHILE}[1]{\textbf{while}$\:\:${#1}$\:\:$\textbf{do}}
\newcommand{\dWHILE}[2]{\textbf{while}$\:\:${#1}$\:\:$\textbf{do}$\:\:${#2}}
\newcommand{\uFOR}[1]{\textbf{for}$\:\:${#1}$\:\:$\textbf{do}}
\newcommand{\FOREACH}[1]{\textbf{foreach}$\:\:${#1}$\:\:$\textbf{do}}
\newcommand{\FOR}[2]{\textbf{for}$\:\:${#1}$\:\:$\textbf{in}$\:\:${#2}$\:\:$\textbf{do}}
\newcommand{\dFOR}[3]{\textbf{for}$\:\:${#1}$\:\:$\textbf{in}$\:\:${#2}$\:\:$\textbf{do}$\:\:${#3}}
\newcommand{\IF}[1]{\textbf{if}$\:\:${#1}}
\newcommand{\ELSE}[1]{\textbf{else}$\:\:${#1}}
\newcommand{\uIF}[1]{\textbf{if}$\:\:${#1}$\:\:$\textbf{then}}
\newcommand{\tIF}[2]{\textbf{if}$\:\:${#1}$\:\:$\textbf{then}$\:\:${#2}}
\newcommand{\rIF}[2]{\textbf{if}$\:\:${#1}$\:\:$\textbf{return}$\:\:${#2}}
\newcommand{\tELIF}[2]{\textbf{elif}$\:\:${#1}$\:\:$\textbf{then}$\:\:${#2}}
\newcommand{\uELIF}[1]{\textbf{elif}$\:\:${#1}$\:\:$\textbf{then}}
\newcommand{\rTHEN}[1]{{#1}$\:\:$\textbf{then}}
\newcommand{\RETURN}[1]{\textbf{return}$\:\:${#1}}
\newcommand{\BREAK}[1]{\textbf{break}$\:\:${#1}}
\newcommand{\CONTINUE}[1]{\textbf{continue}$\:\:${#1}}
\begin{abstract}
This paper presents a precursory yet novel approach to the question answering task using structural decomposition.
Our system first generates linguistic structures such as syntactic and semantic trees from text, decomposes them into multiple fields, then indexes the terms in each field.
For each question, it decomposes the question into multiple fields, measures the relevance score of each field to the indexed ones, then ranks all documents by their relevance scores and weights associated with the fields, where the weights are learned through statistical modeling.
Our final model gives an absolute improvement of over 40\% to the baseline approach using simple search for detecting documents containing answers.
\end{abstract}
\section{Introduction}
\label{sec:introduction}

Towards machine reading, question answering has recently gained lots of interest among researchers from both natural language processing~\cite{moschitti:11a,yih:13a,hixon:15a} and information retrieval~\cite{schiffman:07a,kolomiyets:11a}.
People from these two research fields, NLP and IR, have shown tremendous progress on question answering, yet only few efforts have been made to adapt technologies from both sides.
The NLP side often tackles the task by analyzing linguistic aspects, whereas the IR side tackles it by searching likely patterns.

While these two approaches perform well individually, more sophisticated solutions are needed to handle a wide range of questions.
By considering linguistic structures such as syntactic and semantic trees, QA systems can infer deeper meaning of the context and handle more complex questions.
However, extracting answers from these structures through either graph matching or predicate logic is not necessarily scalable when the size of the context is large.
On the other hand, searching patterns is scalable for large data, especially when coupled with indexing, although it does not always concern with the actual meaning of the context.

We present a multi-field weighted indexing approach for question answering that combines good aspects of both NLP and IR.
We begin by describing how linguistic structures are decomposed into multiple fields (Section~\ref{ssec:fields}), and explain how the decomposed fields are used to rank documents containing answers through statistical learning (Sections~\ref{ssec:answer_ranking} and \ref{ssec:training}).
We evaluate our approach to 8 types of questions; our final model shows significant improvement over the baseline model using simple search (Section~\ref{sec:experiments}).


\section{Related Work}
\label{sec:related-work}

\begin{figure*}[ht]
    \centering
    \includegraphics[scale=0.59]{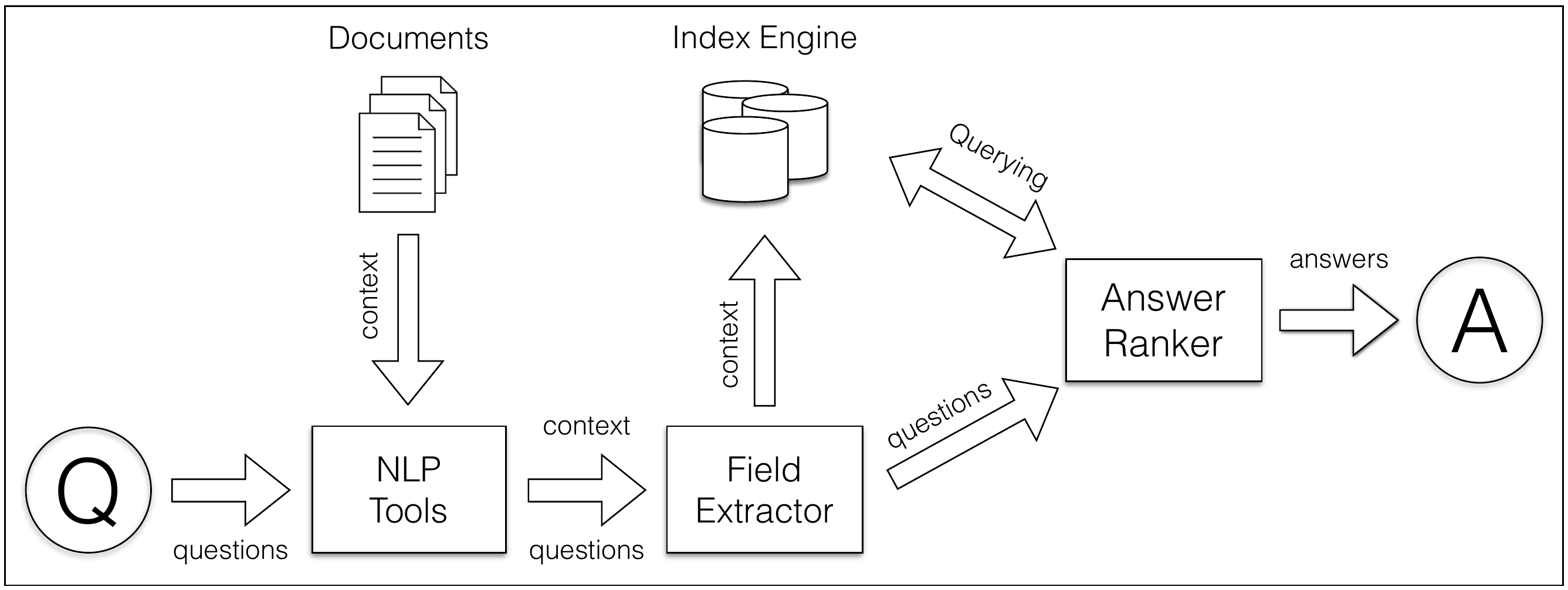}
    \caption{The overall framework of our question answering system.}
    \label{fig:fig1}
\end{figure*}

\newcite{shen:07a} assessed the contribution of semantic roles to factoid question answering and showed promising results.
\newcite{pizzato:08a} proposed a question prediction language model providing rich information and achieved improved speed and accuracy.
Although related, our work is distinguished from theirs because we consider multiple fields whereas the others consider only one field representing semantic roles.
\newcite{ferrucci:10a} presented IBM Watson taking a hybrid approach between NLP and IR, and advanced the question answering task to another level.

\newcite{fader:13a} proposed a paraphrase-driven perceptron learning approach using a seed lexicon.
Our learning process is similar; however, it is distinguished in a way that we learn weights for individual fields instead of lexicons.
\newcite{yih:14a} introduced a semantic parsing framework for open domain question answering, which used convolutional neural networks for measuring similarities between decomposed entities. \newcite{weston15a} presented the Memory Networks models designed to memorize information about known objects and actors.
Our work is related to the this work; however, memory networks are designed to store and manipulate information about specific types of objects while our framework is generalizable to any type of objects induced from the context.

\section{Approach}
\label{sec:approach}

\subsection{Overall framework}

Figure~\ref{fig:fig1} shows the overall framework.
Our system is designed in a modular architectural way, so any further extension of fields can be easily integrated.
The system takes input documents, generates linguistic structures using NLP tools, decomposes them into multiple fields, and indexes those fields.
Questions are processed in the same way.
To answer a question, the system queries the index for each field extracted from the question and measures the relevance score.
All documents are ranked with respect to the relevance scores and their weights associated with the fields, and the document with the highest score is selected as the answer. 

\subsection{Modules}

Our system consists of several modules closely connected together providing a fully working solution for the question answering selection task.

\subsubsection{Documents and questions}

Documents provide the context where the questions find their answers from.
Each document can contain one or more sentences, in which answers for coming questions are annotated for training.
Documents may simply be Wikipedia articles, news articles, fictional stories, etc.
Questions are treated as regular documents containing only one sentence.

\subsubsection{NLP tools}

For the generation of syntactic and semantic structures, we used the part-of-speech tagger~\cite{choi:12a}, the dependency parser~\cite{choi2013transition}, the semantic role labeler~\cite{choi:11b}, and the coreference resolution tool in ClearNLP\footnote{\url{http://www.clearnlp.com}}.
Ensuring good and robust accuracy for these NLP tools is important because all the following modules depend on their output.

\subsubsection{Field extractor}

The field extractor takes the linguistic structures from the NLP tools and decomposes them into multiple fields (Section~\ref{ssec:fields}).
All fields extracted from the documents are passed to the index engine, whereas fields extracted from the questions are sent directly to the answer ranker module.

\subsubsection{Index engine}

The index engine is a search server that receives a list of fields decomposed by the field extractor, indexes terms in the fields, and responses to the queries generated from questions with their relevance scores.
We used Elastic Search\footnote{\url{https://www.elastic.co}}, as it provides a distributed, multi-tenancy-capable search.

\subsubsection{Answer ranker}

The answer ranker takes the decomposed fields extracted from a question, converts them into queries, and builds a matrix of documents with their relevance scores across all fields through the index engine (Section~\ref{ssec:answer_ranking}).
It also uses different weights for individual fields trained by statistical modeling (Section~\ref{ssec:training}).

\subsection{Structural decomposition}
\label{ssec:fields}


\begin{figure}[ht]
    \centering
    \includegraphics[scale=0.63]{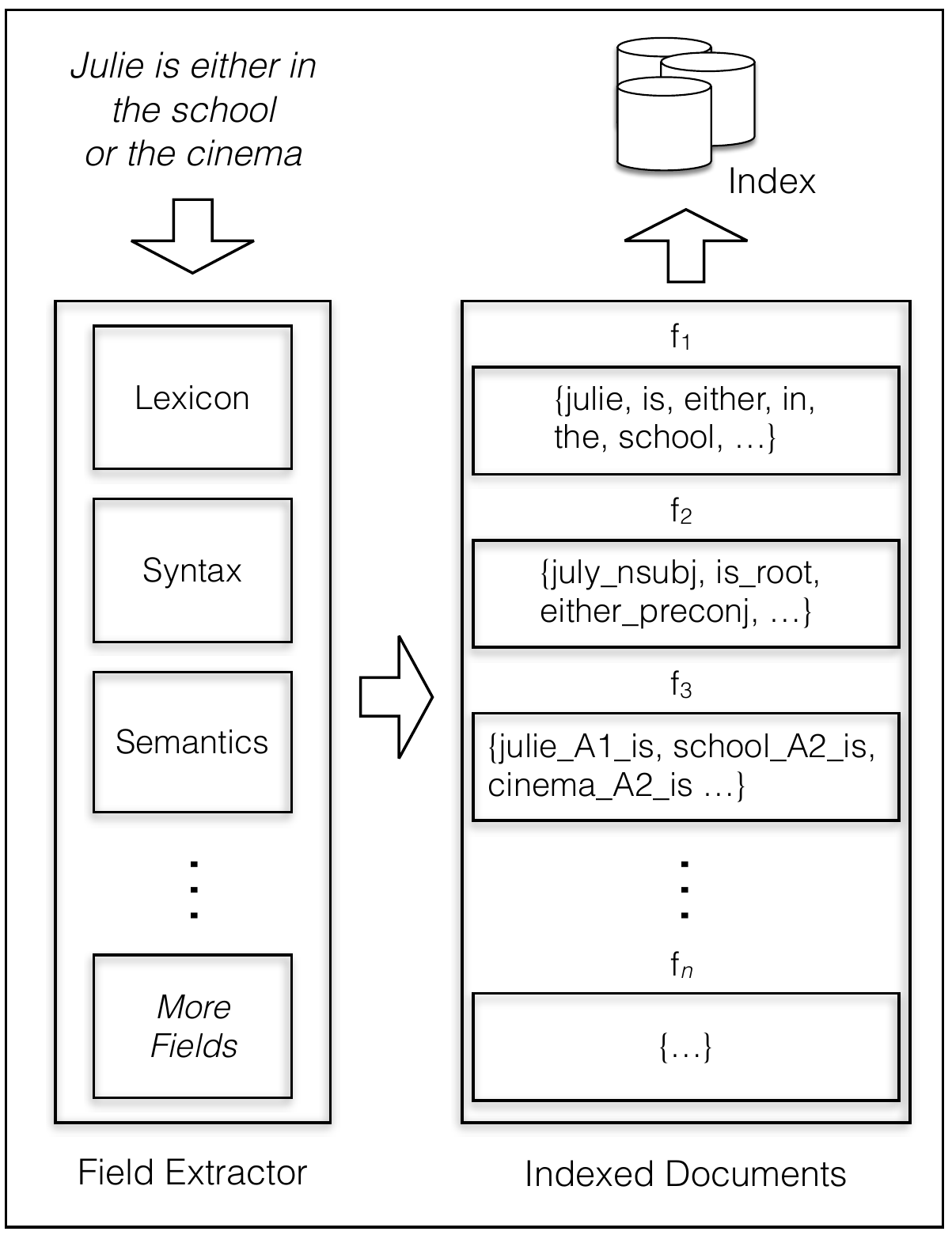}
    \caption{The flow of the sentence, \textit{Julie is either in the school or the cinema}, through our system.}
    \label{fig:fig2}
\end{figure}

\noindent Each sentence is represented by the index engine as a document with multiple fields grouped into categories.
Figure~\ref{fig:fig2} shows an example of how the sentence is decomposed into multiple fields consisting of syntactic and semantic structures.
Due to the extensible nature of our field extractor, additional groups and fields can be easily integrated.
Currently, our system supports 24 fields grouped into the following three categories:

\begin{itemize}
\item Lexical fields (e.g., word-forms, lemmas).
\item Syntactic fields (e.g., dependency labels).
\item Semantic fields (e.g., semantic roles, distances between predicates).
\end{itemize}

\subsection{Answer ranking}
\label{ssec:answer_ranking}

When a question $q$ is asked, it is decomposed into the $n$-number of fields.
Each field is transformed into a query where certain words are replaced with wildcards (e.g., \{\textit{where}\_a1, \textit{is}\_pred, \textit{she}\_a2\} $\rightarrow$ \{*\_a1 \textit{is}\_pred \textit{she}\_a2\}).
Then, the relevance score $r$ is measured between each field in the question and the same field in each document $d^t \in D$ by the index engine.\footnote{ We set the elastic search results limit to 20.}
The product of the relevance scores and individual weights for all fields are summed, and the document $\hat{d}$ with the highest score $f$ is taken as the answer.
Note that in our dataset, each document contains only one sentence so that retrieving a document is equivalent to retrieving a sentence.
The following equations describe how the document $\hat{d}$ is selected by measuring the overall score $f(q,d^t)$ using the relevance scores $r(q_i,d_i^t)$ and the weights $\lambda_i$.

\begin{align*}
\hat{d} &= \arg\max_{d^t \in D} f(q,d^t)\\
f(q,d^t) &= \sum_{i=1}^{n} \lambda_i \cdot r(q_i, d_i^t)\\
r(q_i, d_i^t) &= \sum_{v\in q_i \cap d_i^t} \mathrm{tf}_i^t(v) \cdot \mathrm{idf}_i(v) \cdot \mathrm{norm}_i^t(v)\\
\end{align*}

\subsection{Training weights for individual fields}
\label{ssec:training}

Algorithm~\ref{alg:training} shows how the weights for all fields are learned during training.
We adapt the averaged perceptron algorithm, which has been widely used for many NLP tasks.
All the weights $\vec{\lambda}$ are initialized to 1.
For each question $q \in Q$, it predicts the document $\hat{d}$ that most likely contains the answer.
If $\hat{d}$ is incorrect, then it compares the relevance score $r$ between $(q,\hat{d})$ and $(q,d)$ for each field, and updates the weight accordingly, where $d$ is the true document from the oracle.
This procedure is repeated multiple times through iterations.
Finally, the algorithm returns the averaged weights, where each dimension represents the weight for each field.

\begin{algorithm}[htbp!]
\small
\setcounter{LINE_NUM}{0}
\vspace{1ex}
\begin{tabular}{rp{13cm}}
\bf Input:  & $D\mathrm{: document\;set}, Q\mathrm{: question\;set}$.\\
            & $M$: max-number of iterations, $\alpha$: learning rate.\\
\bf Output: & The averaged weight vector.\\
\end{tabular}
\vspace{1ex}

\begin{tabular}{p{0.4cm}l}
$\:\:$\LN & $\vec{\lambda}\leftarrow 1;\:\:\vec{\lambda'}\leftarrow 0$\\
$\:\:$\LN & \uFOR{$iter \in [1, M]$}\\
$\:\:$\LN & \TAB\FOREACH{$q \in Q$}\\
$\:\:$\LN & \TAB\TAB$\hat{d} = \arg\max_{d^t \in D} f(q,d^t)$\\
$\:\:$\LN & \TAB\TAB\uIF{$\hat{d} \neq d$}\TAB\TAB\TAB\TAB\TAB\TAB\TAB\:\:\# $d$ is the oracle\\
$\:\:$\LN & \TAB\TAB\TAB\FOREACH{$i \in [1,n]$}\TAB\TAB\TAB\# for each field\\
$\:\:$\LN & \TAB\TAB\TAB\TAB$\delta \leftarrow \alpha\cdot\mathrm{sign}[r(q_i, d_i) - r(q_i, \hat{d}_i)]$\\
$\:\:$\LN & \TAB\TAB\TAB\TAB$\lambda_i \leftarrow \lambda_i + \delta $\\
$\:\:$\LN & \TAB\TAB$\vec{\lambda'} \leftarrow \vec{\lambda'} + \vec{\lambda}$\\
$\:\:$\LN & \RETURN{$\vec{\lambda'}\cdot\frac{1}{M * |Q|}$}\\
\end{tabular}
\caption{Averaged perceptron training.}
\label{alg:training}
\end{algorithm}

\noindent All hyper-parameters were optimized on the development sets and evaluated on the test sets.
For our experiments, we used the following hyper-parameters: $M = 40, \alpha = 0.002$.

\begin{table*}[ht]
\small\centering
\begin{tabular}{l||cc|cc||cc|cc||cc|cc}
\multirow{3}{*}{Type} & \multicolumn{4}{c||}{Lexical} & \multicolumn{4}{c||}{Lexical + Syntax} & \multicolumn{4}{c}{Lexical + Syntax + Semantics} \\
 & \multicolumn{2}{c|}{$\lambda = 1$} & \multicolumn{2}{c||}{$\lambda$ is learned} & \multicolumn{2}{c|}{$\lambda = 1$} & \multicolumn{2}{c||}{$\lambda$ is learned} & \multicolumn{2}{c|}{$\lambda = 1$} & \multicolumn{2}{c}{$\lambda$ is learned}\\
 & MAP & MRR & MAP & MRR & MAP & MRR & MAP & MRR & MAP & MRR & MAP & MRR\\
\hline\hline
1 (qa1) & 39.62 & 61.73 & 39.62 & 61.73 & 29.90 & 48.05 & 40.50 & 61.47 & 72.60 & 85.07 & \textbf{100.0} & \textbf{100.0}\\
2 (qa4) & 62.90 & 81.45 & 62.90 & 81.45 & 64.00 & 82.00 & 64.00 & 82.00 & 55.70 & 77.85 & \textbf{64.10} & \textbf{82.05}\\
3 (qa5) & 37.10 & 54.00 & 38.20 & 54.70 & 48.00 & 62.15 & 48.40 & 62.25 & 72.60 & 82.65 & \textbf{94.20} & \textbf{96.33}\\
4 (qa6) & 64.00 & 75.07 & 64.00 & 75.07 & 65.80 & 78.47 & 66.10 & 78.53 & 78.20 & 88.33 & \textbf{89.30} & \textbf{94.27}\\
5 (qa9) & 47.90 & 63.50 & 48.10 & 63.62 & 47.90 & 63.67 & 50.50 & 65.47 & 53.90 & 67.88 & \textbf{94.40} & \textbf{96.72}\\
6 (qa10)& 47.80 & 63.78 & 47.90 & 63.92 & 49.20 & 65.52 & 50.20 & 66.33 & 57.60 & 70.68 & \textbf{96.90} & \textbf{98.23}\\
7 (qa12)& 19.20 & 38.68 & 19.20 & 38.68 & 25.10 & 40.83 & 31.90 & 49.82 & 55.00 & 70.60 & \textbf{99.60} & \textbf{99.80}\\
8 (qa20)& 37.10 & 51.82 & 37.10 & 51.82 & 31.40 & 42.00 & 35.70 & 44.22 & 31.20 & 46.50 & \textbf{42.80} & \textbf{56.32}\\
\hline\hline
\multicolumn{1}{c||}{Avg.} & 44.45 & 61.25 & 44.63 & 61.37 & 45.16 & 60.34 & 48.41 & 63.76 & 59.60 & 73.70 & \textbf{85.16} & \textbf{90.47}\\
\end{tabular}
\caption{Results from our question-answering system on 8 types of questions in the bAbI tasks.}
\label{tbl:results}
\end{table*}

\section{Experiments}
\label{sec:experiments}

\subsection{Data and evaluation metrics}

Our approach is evaluated on a subset of the bAbI tasks~\cite{weston15a}.
The original data contains 20 tasks, where each task represents a different kind of question answering challenge.
We select 8 tasks, in which answer for a single question is located within a single sentence.
For consistency and replicability, we follow the same training, development, and evaluation set splits as provided, where every set contains 1,000 questions.

For the evaluation metrics, we use mean average precision (\textsc{map}) and mean reciprocal rank (\textsc{mrr}) of the top-3 predictions.
The mean average precision is measured by counting the number of questions, for which sentences containing the answers are correctly selected as the best predictions.
The reciprocal rank of a query response is the multiplicative inverse of the rank of the first correct answer. Mean reciprocal rank is the average of the reciprocal ranks of all question queries.

\subsection{Evaluation}

Table~\ref{tbl:results} shows the results from our system on different types of questions.
The \textsc{map} and \textsc{mrr} show clear correlation with respect to the number of active fields.
For the majority of tasks, using only the lexical fields does not perform well.
The fictional stories included in this data often contain multiple occurrences of the same lexicons, and the lexical fields alone are not able to select the correct answer.
Significantly lower accuracy for the last task is due to a fact that besides an answer is located within a single sentence, multiple passages for the single question are required to correctly locate the sentence with the answers.
Lexical fields coupled with only syntactic fields do not perform much better.
It may be due to a fact that the syntactic fields containing ordinary dependency labels do not provide sufficient context-wise information so that they do not generate enough features for statistical learning to capture specific characteristic of the context.
The significant improvement, however, is reached when the semantics fields are added as they provide deeper understanding of the context.

Not that this data set has also been used for evaluating the Memory Networks approach to question answering~\cite{weston15a}.
The authors achieved high accuracy, reaching 100\% in several tasks; however, our work still finds its own value because our approach is completely data-driven such that it can be easily adapted or extended to other types of questions.
As a matter of fact, we are using the same system for all tasks with different trained models, yet still able to achieve high accuracy for most tasks we evaluate on.
\section{Conclusion}
\label{sec:conclusion}

This paper presents a multi-field weighted indexing approach for question answering.
Our system decomposes linguistic structures into multiple fields, indexes terms of individual fields, and retrieves the documents containing the answers with respect to the relevance scores weighted differently.
We observe significant improvement as we add more semantic fields and apply averaged perceptron learning to statistically designate weights for the fields.

In the future, we plan to extend our work by integrating additional layers of fields (e.g., Freebase, WordNet).
Furthermore, we plan to improve our NLP tools to enable even deeper understanding of the context for more complex question answering.

\bibliographystyle{acl}
\bibliography{references}

\end{document}